\documentclass{article}

\usepackage[preprint, nonatbib]{neurips_2023}

\usepackage[utf8]{inputenc} % allow utf-8 input
\usepackage[T1]{fontenc}    % use 8-bit T1 fonts
\usepackage{hyperref}       % hyperlinks
\usepackage{url}            % simple URL typesetting
\usepackage{booktabs}       % professional-quality tables
\usepackage{amsfonts}       % blackboard math symbols
\usepackage{nicefrac}       % compact symbols for 1/2, etc.
\usepackage{microtype}      % microtypography
\usepackage{xcolor}         % colors

\usepackage{eccvabbrv}

\usepackage{graphicx}
\usepackage{booktabs}
\usepackage{epsfig}
\usepackage{amsmath}
\usepackage{amssymb}
\usepackage{multirow}
\usepackage{overpic}
\usepackage{placeins}
\usepackage{dashrule}
\usepackage{blindtext, rotating}
\usepackage{color, colortbl}
\usepackage[misc]{ifsym}
\usepackage{subcaption}
\usepackage[hang,flushmargin]{footmisc}
\usepackage{listings}

\usepackage[accsupp]{axessibility}  % Improves PDF readability for those with disabilities.
\usepackage{hyperref}

\newcommand{\authornote}[1]{{%
  \let\thempfn\relax% Remove footnote number printing mechanism
  \footnotetext[0]{#1}% Print footnote text
}}

\newcommand{\TP}{\mathit{TP}}
\newcommand{\FN}{\mathit{FN}}
\newcommand{\FP}{\mathit{FP}}

\begin{document}

\title{Fully Sparse 3D Occupancy Prediction} 


\author{
Haisong Liu$^{1,2*}$,
Yang Chen$^{1*}$,
Haiguang Wang$^{1}$,
Zetong Yang$^2$,
\textbf{Tianyu Li}$^{2}$, \\
\textbf{Jia Zeng}$^{2}$,
\textbf{Li Chen}$^{2}$,
\textbf{Hongyang Li}$^{2}$,
\textbf{Limin Wang}$^{1,2,}$\textsuperscript{\Letter}
\\ [0.15cm]
$^1$Nanjing University~~~~$^2$Shanghai AI Lab~~~~
\\ [0.25cm]
\textbf{\url{https://github.com/MCG-NJU/SparseOcc}}
}

% \authorrunning{Liu \etal}

% \institute{State Key Laboratory for Novel Software Technology, Nanjing University \and Shanghai AI Lab}

\maketitle

\authornote{*: Equal contribution.}
\authornote{\Letter: Corresponding author.}

\begin{abstract}
  Occupancy prediction plays a pivotal role in autonomous driving. Previous methods typically construct dense 3D volumes, neglecting the inherent sparsity of the scene and suffering from high computational costs. To bridge the gap, we introduce a novel fully sparse occupancy network, termed SparseOcc. SparseOcc initially reconstructs a sparse 3D representation from camera-only inputs and subsequently predicts semantic/instance occupancy from the 3D sparse representation by sparse queries. A mask-guided sparse sampling is designed to enable sparse queries to interact with 2D features in a fully sparse manner, thereby circumventing costly dense features or global attention. Additionally, we design a thoughtful ray-based evaluation metric, namely RayIoU, to solve the inconsistency penalty along the depth axis raised in traditional voxel-level mIoU criteria. SparseOcc demonstrates its effectiveness by achieving a RayIoU of 34.0, while maintaining a real-time inference speed of 17.3 FPS, with 7 history frames inputs. By incorporating more preceding frames to 15, SparseOcc continuously improves its performance to 35.1 RayIoU without bells and whistles.
  % \keywords{3D Occupancy Estimation \and Semantic Scene Completion \and 3D Reconstruction \and Autonomous Driving}
\end{abstract}

\section{Introduction}
\label{sec:intro}

Vision-centric 3D occupancy prediction \cite{tesla_ai_day} focuses on partitioning 3D scenes into structured grids from visual images. Each grid is assigned a label indicating if it is occupied or not. This task offers more geometric details than 3D object detection and produces an alternative representation to LiDAR-based perception \cite{centerpoint,pointpillars,3dssd,std,3dman,camliflow,camliraft}.

\begin{figure*}[t]
    \centering
    \subfloat[Overview of SparseOcc]{\includegraphics[width=0.56\linewidth]{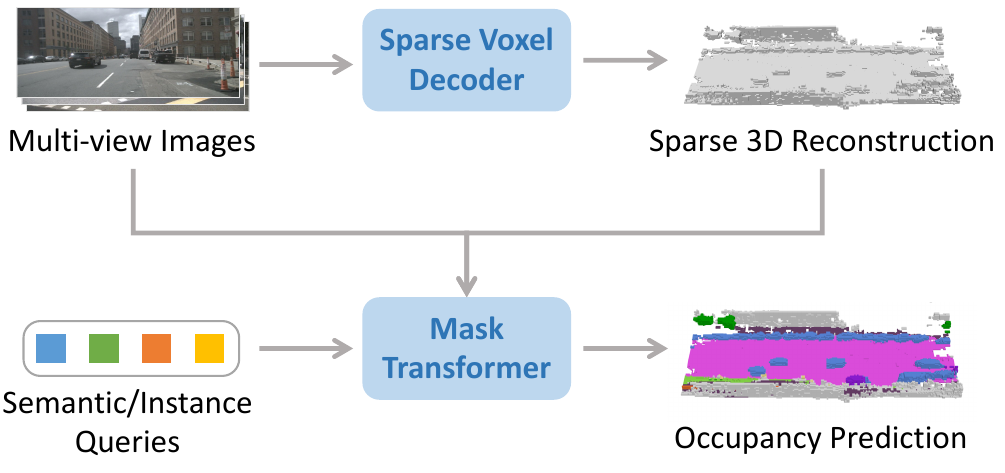}}%
    \hfill
    \subfloat[Performance Comparison]{\includegraphics[width=0.37\linewidth]{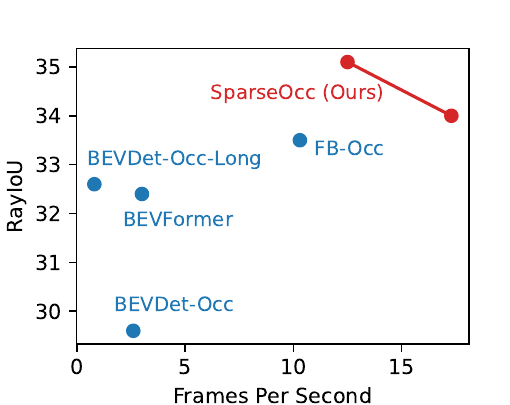}}%
    \caption{(a) SparseOcc reconstructs a sparse 3D representation from camera-only inputs by a sparse voxel decoder, and then estimates the mask and label of each segment via a set of sparse queries. (b) Performance comparison on the validation split of Occ3D-nuScenes. FPS is measured on a Tesla A100 with the PyTorch \texttt{fp32} backend.}
\label{fig:intro}
\end{figure*}

Existing methods \cite{bevformer,bevdet,surroundocc,occnet,fbocc} typically construct dense 3D features yet suffer from computational overhead (\eg, $2 \sim 3$ FPS on the Tesla A100 GPU). However, dense representations are not necessary for occupancy prediction. We statistic the geometry sparsity and find that more than 90\% of the voxels are empty. This manifests a large room in occupancy prediction acceleration by exploiting the sparsity. Some works \cite{voxformer,tpvformer} explore the sparsity of 3D scenes, but they still rely on sparse-to-dense modules for dense predictions. This inspires us to seek a pure sparse occupancy network without any dense design.

In this paper, we propose SparseOcc, the first fully sparse occupancy network. As depicted in Fig.~\ref{fig:intro} (a), SparseOcc includes two steps.
First, it leverages a \textit{sparse voxel decoder} to reconstruct the sparse geometry of a scene in a coarse-to-fine manner. This only models non-free regions, saving computational costs significantly.
Second, we design a \textit{mask transformer} with sparse semantic/instance queries to predict masks and labels of segments from the sparse space.
The mask transformer not only improves performance on semantic occupancy but also paves the way for panoptic occupancy.
A \textit{mask-guided sparse sampling} is designed to achieve sparse cross-attention in the mask transformer.
As such, our SparseOcc fully exploits the sparse property and gets rid of any dense design like dense 3D features, sparse-to-dense modules, and global attention.

Besides, we notice flaws in popular voxel-level mean Intersection-over-Union (mIoU) metrics for occupancy evaluation and further design a ray-level evaluation, RayIoU, as the solution.
The mIoU criterion is an ill-posed formulation given the ambiguous labeling of unscanned voxels. Previous methods\cite{occ3d} relieve this issue by only evaluating observed areas but raise extra issues in inconsistency penalty along depths.
Instead, RayIoU addresses the two aforementioned issues simultaneously. It evaluates predicted 3D occupancy volume by retrieving depth and category predictions of designated rays. To be specific, RayIoU casts query rays into predicted 3D volumes and decides true positive predictions as the ray with the correct distance and class of its first touched occupied voxel grid. This formulates a more fair and reasonable criterion.

Thanks to the sparsity design, SparseOcc achieves 34.0 RayIoU on Occ3D-nuScenes \cite{occ3d}, while maintaining a real-time inference speed of 17.3 FPS (Tesla A100, PyTorch \texttt{fp32} backend), with 7 history frames inputs. By incorporating more preceding frames to 15, SparseOcc continuously improves its performance to 35.1 RayIoU, achieving state-of-the-art performance without bells and whistles. The comparison between SparseOcc with previous methods in terms of performance and efficiency is shown in Fig.~\ref{fig:intro} (b).

We summarize our contributions as follows:
\begin{enumerate}
  \item We propose SparseOcc, the first fully sparse occupancy network without any time-consuming dense designs. It achieves 34.0 RayIoU on Occ3D-nuScenes benchmark with an real-time inference speed of 17.3 FPS.
  \item We present RayIoU, a ray-wise criterion for occupancy evaluation. By querying rays to 3D volume, it solves the ambiguous penalty issue for unscanned free voxels and the inconsistent depth penalty issue in the mIoU metric.
\end{enumerate}

\section{Related Work}

\paragraph{Camera-based 3D Occupancy Prediction.} The occupancy network is originally proposed by Mescheder \etal~\cite{occupancy_network, convolutional_occupancy_networks}, focusing on continuous object representations in 3D space.
Recent variations~\cite{tesla_ai_day, monoscene, occnet, occ3d, openoccupancy, panoocc, simpleocc, yang2023vidar} mostly draw inspiration from Bird's Eye View (BEV) perception \cite{li2023bevsurvey,li2023opensourced,bevformer,bevdet,bevdet4d,gapretrain,vcd,detr3d,petr,petrv2,sparse4d,sparsebev,streampetr,eq2022yang} and predicts voxel-level semantic information from image inputs.
For instance, MonoScene~\cite{monoscene} estimates occupancy through a 2D and a 3D UNet~\cite{unet} connected by a sight projection module.
SurroundOcc \cite{surroundocc} proposes a coarse-to-fine architecture. However, the large number of voxel queries is computationally heavy.
TPVFormer~\cite{tpvformer} proposes tri-perspective view representations to supplement vertical structural information, but this inevitably leads to information loss.
VoxFormer~\cite{voxformer} initializes sparse queries based on monocular depth prediction. Nevertheless, VoxFormer is not fully sparse as it still requires a sparse-to-dense MAE~\cite{mae} module to complete the scene.
Some methods emerged in the CVPR 2023 occupancy challenge~\cite{fbocc, renderocc, multiscaleocc}, but none of them exploits a fully sparse design.
In this paper, we make the first step to explore the fully sparse architecture for 3D occupancy prediction from camera-only inputs.

\paragraph{Sparse Architectures for 3D Vision.} Sparse architectures find widespread adoption in LiDAR-based reconstruction \cite{takikawa2021neural} and perception \cite{Minkowski4d, centerpoint, 3dssd, std}, leveraging the inherent sparsity of point clouds. However, when it comes to vision-to-3D tasks, a direct adaptation is not feasible due to the absence of point cloud inputs. A prior work, SparseBEV \cite{sparsebev}, proposes a fully sparse architecture for camera-based 3D object detection. Nevertheless, directly adapting this approach is non-trivial because 3D object detection focuses on a sparse set of objects, whereas 3D occupancy requires dense predictions for each voxel. Consequently, designing a fully sparse architecture for 3D occupancy prediction remains a challenging task.

\paragraph{End-to-end 3D Reconstruction from Posed Images.} As a related task to 3D occupancy prediction, 3D reconstruction recovers the 3D geometry from multiple posed images. Recent methods focus on more compact and efficient end-to-end 3D reconstruction pipelines \cite{atlas,neuralrecon,transformerfusion,vortx,cvrecon}. Atlas~\cite{atlas} extracts features from multi-view input images and maps them to 3D space to construct the truncated signed distance function~\cite{sdf}. NeuralRecon~\cite{neuralrecon} directly reconstructs local surfaces as sparse TSDF volumes and uses a GRU-based TSDF fusion module to fuse features from previous fragments. VoRTX~\cite{vortx} utilizes transformers to address occlusion issues in multi-view images. % CVRecon~\cite{cvrecon} takes a different approach by avoiding occlusion-sensitive perspective mappings of 2D to 3D and directly uses cost volumes to establish 3D features from 2D image features. 

\paragraph{Mask Transformer.} Recently, unified segmentation models have been widely studied to handle semantic and instance segmentation concurrently. Cheng \etal first propose MaskFormer~\cite{maskformer} for unified segmentation in terms of model architecture, loss functions, and training strategies. Mask2Former~\cite{mask2former} then introduces masked attention, with restricted receptive fields on instance masks, for better performance. Later on, Mask3D~\cite{mask3d} successfully extends the mask transformer for point cloud segmentation with state-of-the-art performance. OpenMask3D~\cite{openmask3d} further achieves the open-vocabulary 3D instance segmentation task and proposes a model for zero-shot 3D segmentation.

\begin{figure*}[t!]
  %\centering
  \includegraphics[width=\linewidth]{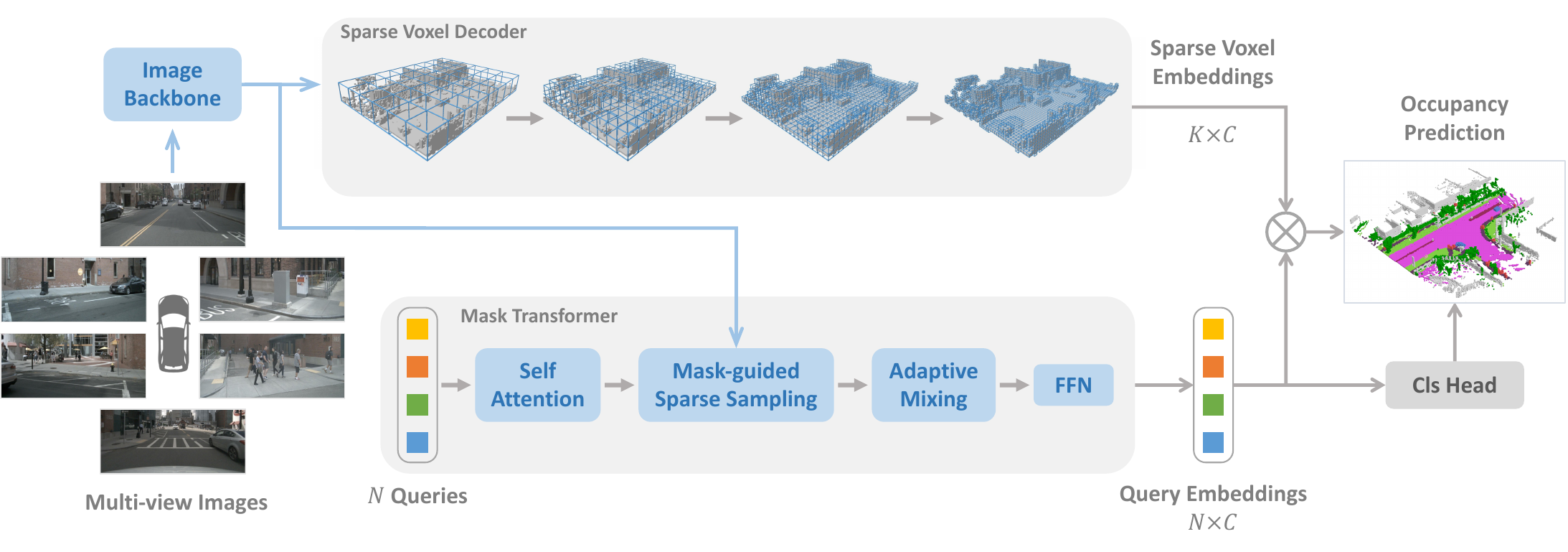}
  \caption{SparseOcc is a fully sparse architecture since it neither relies on dense 3D feature, nor has sparse-to-dense and global attention operations. The sparse voxel decoder reconstructs the sparse geometry of the scene, consisting of $K$ voxels ($K \ll W \times H \times D$). The mask transformer then uses $N$ sparse queries to predict the mask and label of each segment. SparseOcc can be easily extended to panoptic occupancy by replacing the semantic queries with instance queries.}
  \label{fig:arch}
\end{figure*}

\section{SparseOcc}

SparseOcc is a vision-centric occupancy model that only requires camera inputs. As shown in Fig. \ref{fig:arch}, SparseOcc has three modules: an image encoder consisting of an image backbone and FPN \cite{fpn} to extract 2D features from multi-view images; a sparse voxel decoder (Sec. \ref{sec:sparse-voxel-decoder}) to predict sparse class-agnostic 3D occupancy with correlated embeddings from the image features; a mask transformer decoder (Sec \ref{sec:mask-transformer}) to distinguish semantics and instances in the sparse 3D space.

\subsection{Sparse Voxel Decoder}
\label{sec:sparse-voxel-decoder}

Since 3D occupancy ground truth \cite{occ3d,occnet,surroundocc,openoccupancy} is a dense volume with dimensions $W \times H \times D$ (\eg, 200$\times$200$\times$16), existing methods typically build a dense 3D feature of shape $W \times H \times D \times C$, but suffer from computational overhead. 
In this paper, we argue that such dense representation is not necessary for occupancy prediction. As in our statistics, we find that over 90\% of the voxels in the scene are free. This motivates us to explore a sparse 3D representation that only models the non-free areas of the scene, thereby saving computational resources.

\begin{figure*}[t]
  %\centering
  \includegraphics[width=\linewidth]{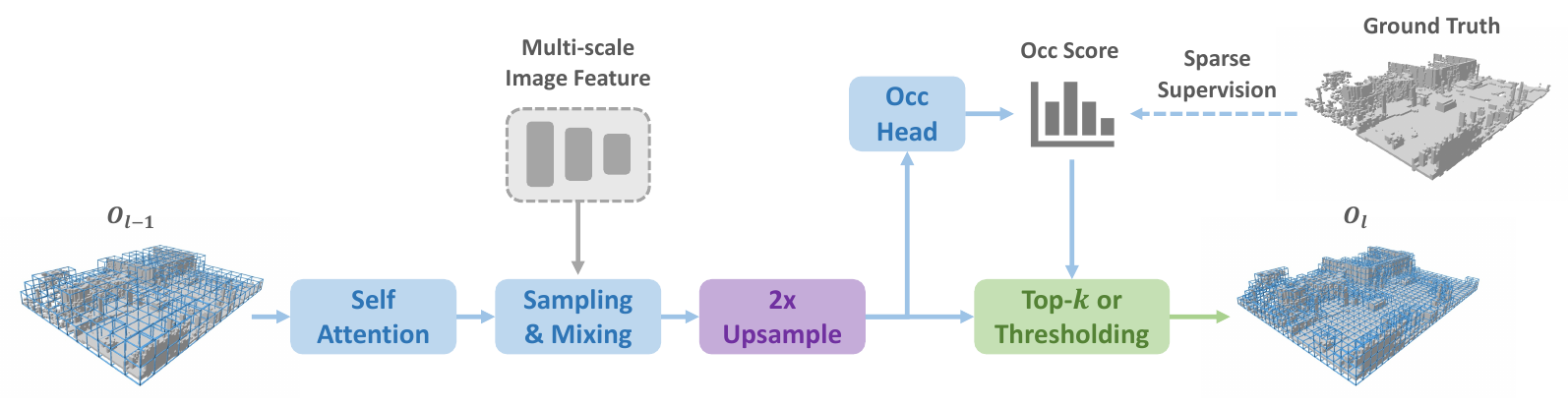}
  \caption{The sparse voxel decoder employs a coarse-to-fine pipeline with three layers. Within each layer, we utilize a transformer-like architecture for 3D-2D interaction. At the end of every layer, the voxel resolution is upsampled by a factor of 2$\times$, and probabilities of voxel occupancy are estimated.}
  \label{fig:sparse-voxel-decoder-arch}
\end{figure*}

\paragraph{Overall architecture.} Our designed sparse voxel decoder is shown in Fig. \ref{fig:sparse-voxel-decoder-arch}. In general, it follows a coarse-to-fine structure but only models the non-free regions.
The decoder starts from a set of coarse voxel queries equally distributed in the 3D space (\eg, 25$\times$25).
In each layer, we first upsample each voxel by 2$\times$, \eg, a voxel with size $d$ will be upsampled into 8 voxels with size $\frac{d}{2}$. Next, we estimate an occupancy score for each voxel and conduct pruning to remove useless voxel grids. Here we have two approaches for pruning: one is based on a threshold (\eg, only keeps score $>$ 0.5); the other is by top-$k$ selection. In our implementation, we simply keep voxels with top-$k$ occupancy scores for training efficiency. $k$ is a dataset-related parameter, obtained by counting the maximum number of non-free voxels in each sample at different resolutions. The voxel tokens after pruning will serve as the input for the next layer.

\paragraph{Detailed design.} Within each layer, we use a transformer-like \cite{transformer} architecture to handle voxel queries.
The concrete architecture is inspired by SparseBEV \cite{sparsebev}, a detection method using a sparse scheme.
To be specific, in layer $l$ with $K_{l-1}$ voxel queries described by 3D locations and a $C$-dim content vector, we first use self-attention to aggregate local and global features for those query voxels.
Then, a linear layer is used to generate 3D sampling offsets $\{(\Delta x_i, \Delta y_i, \Delta z_i)\}$ for each voxel query from the associated content vector.
These sampling offsets are utilized to transform voxel queries to obtain reference points in global coordinates.
We finally project those sampled reference points to multi-view image space for integrating image features by adaptive mixing \cite{adamixer, mlpmixer, dynamic_conv}.
In summary, our approach differs from SparseBEV by shifting the query formulation from pillars to 3D voxels. Other components such as self attention, adaptive sampling and mixing are directly borrowed.

\paragraph{Temporal modeling.} Previous dense occupancy methods \cite{bevformer, bevdet} typically warp the history BEV/3D feature to the current timestamp, and use deformable attention \cite{deformabledetr} or 3D convolutions to fuse temporal information. However, this approach is not directly applicable in our case due to the sparse nature of our 3D features. To handle this, we leverage the flexibility of the aforementioned global sampled reference points by warping them to previous timestamps to sample history multi-view image features. The sampled multi-frame features are stacked and aggregated by adaptive mixing so as for temporal modeling.

\paragraph{Supervision.} We compute loss for the sparsified voxels from each layer. We use \textit{binary cross entropy} (BCE) loss as the supervision, given that we are reconstructing a class-agnostic sparse occupancy space. Only the kept sparse voxels are supervised, while the discarded regions during pruning in earlier stages are ignored.

Moreover, due to the severe class imbalance, the model can be easily dominated by categories with a large proportion, such as the ground, thereby ignoring other important elements in the scene, such as cars, people, etc. Therefore, voxels belonging to different classes are assigned with different loss weights. For example, voxels belonging to class $c$ are assigned with a loss weight of:
\begin{align}
  w_c = \frac{\sum_{i=1}^C M_i}{M_c},
\end{align}
where $M_i$ is the number of voxels belonging to the $i$-th class in ground truth.

\subsection{Mask Transformer}
\label{sec:mask-transformer}

Our mask transformer is inspired by Mask2Former \cite{mask2former}, which uses $N$ sparse semantic/instance queries decoupled by binary mask queries $\mathbf{Q}_m \in [0, 1]^{N \times K}$ and content vectors $\mathbf{Q}_c \in \mathbb{R}^{N \times C}$. The mask transformer consists of three steps: multi-head self attention (MHSA), mask-guided sparse sampling, and adaptive mixing. MHSA is used for the interaction between different queries as the common practice. Mask-guided sparse sampling and adaptive mixing are responsible for the interaction between queries and 2D image features.

\paragraph{Mask-guided sparse sampling.} A simple baseline of mask transformer is to use the masked cross-attention module in Mask2Former. However, it attends to all positions of the key, with unbearable computations. Here, we design a simple alternative. We first randomly select a set of 3D points within the mask predicted by the previous ($l-1$)-th Transformer decoder layer. 
Then, we project those 3D points to multi-view images and extract their features by bilinear interpolation. Besides, our sparse sampling mechanism makes the temporal modeling easier by simply warping the sampling points (as done in the sparse voxel decoder). %The sampled features are further enhanced by adaptive mixing.

\paragraph{Prediction.} For class prediction, we apply a linear classifier with a sigmoid activation based on the query embeddings $\mathbf{Q}_c$. For mask prediction, the query embeddings are converted to mask embeddings by an MLP. The mask embeddings $\mathbf{M} \in \mathbb{R}^{Q \times C}$ have the same shape as query embeddings $\mathbf{Q}_c$ and are dot-producted with the sparse voxel embeddings $\mathbf{V} \in \mathbb{R}^{K \times C}$ to produce mask predictions. Thus, the prediction space of our mask transformer is constrained to the sparsified 3D space from the sparse voxel decoder, rather than the full 3D scene. The mask predictions will serve as the mask queries $\mathbf{Q}_m$ for the next transformer layer.

\paragraph{Supervision.} The reconstruction result from the sparse voxel decoder may not be reliable, as it may overlook or inaccurately detect certain elements. Thus, supervising the mask transformer presents certain challenges since its predictions are confined within this unreliable space. In cases of missed detection, where some ground truth segments are absent in the predicted sparse occupancy, we opt to discard these segments to prevent confusion. As for inaccurately detected elements, we simply categorize them as an additional ``no object'' category.

\paragraph{Loss Functions.} Following MaskFormer \cite{maskformer}, we match the ground truth with the predictions using Hungarian matching. Focal loss $L_{focal}$ is used for classification, while a combination of DICE loss \cite{vnet} $L_{dice}$ and BCE mask loss $L_{mask}$ is used for mask prediction. Thus, the total loss of SparseOcc is composed of four parts:
\begin{equation}
  L = L_{focal} + L_{mask} + L_{dice} + L_{occ},
\end{equation}
where $L_{occ}$ is the loss of sparse voxel decoder.

\section{Ray-level mIoU}

\subsection{Revisiting the Voxel-level mIoU}

\begin{figure}[t]
  \centering
  \subfloat{\includegraphics[width=\linewidth]{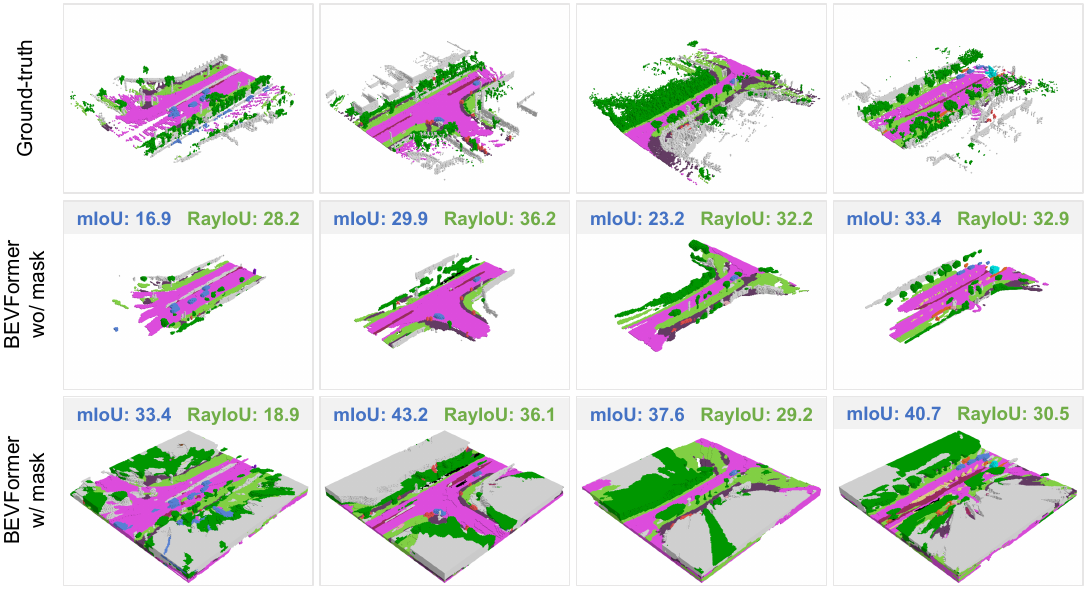}}
  \caption{Visualization of the discrepancy between qualitative and quantitative results. We observe that training existing dense occupancy methods (\eg BEVFormer) with a visible mask results in a thick surface, leading to an unreasonably inflated improvement in the current mIoU metrics. In contrast, our new RayIoU metrics provide a more accurate reflection of model performance.}
  \label{fig:iou-bug-compare}
\end{figure}

The Occ3D dataset~\cite{occ3d}, along with its proposed evaluation metrics, are widely recognized as benchmarks in this field. The ground truth occupancy is reconstructed from LiDAR point clouds, and the mean Intersection over Union (mIoU) at the voxel level is employed to assess performance. 
Due to factors such as distance and occlusion, the accumulated point clouds are not perfect. Some areas unscanned by LiDAR are marked as free, resulting in fragmented instances.
This raises the problem of label inconsistency.
To solve this problem, Occ3D uses a binary \textit{visible mask} that indicates whether a voxel is observed in the current camera view. Only the observed voxels contribute to evaluation.

However, we found that solely calculating mIoU on the observed voxel positions remains vulnerable and can be hacked by \textit{predicting a thicker surface}.
Dense methods (\eg, BEVFormer \cite{bevformer}) can easily achieve this by training with the visible mask.
During training, the area behind the surface lacks supervision, causing the model to fill it with duplicated predictions, resulting in a thicker surface.
As an example, consider BEVFormer, which generates a thick and noisy surface when trained with the visible mask (see Fig.~\ref{fig:iou-bug-compare}). Despite this, its performance exhibits an unreasonably inflated improvement (+5$\sim$15 mIoU) under the current evaluation protocol.

\begin{figure*}[t]
  \includegraphics[width=\linewidth]{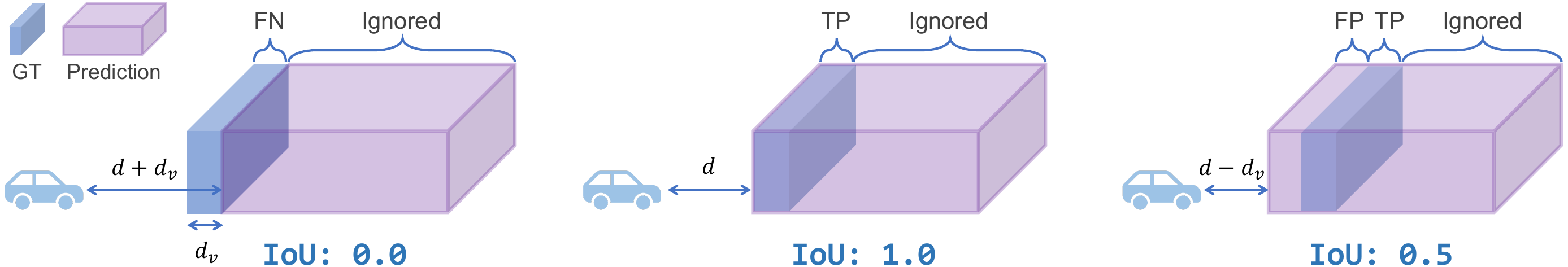}
  \caption{Illustration of inconsistent depth penalties caused by current metrics. Consider a scenario where we have a wall in front of us, with a ground-truth distance of $d$ and a thickness of $d_v$. When the prediction has a thickness of $d_p \gg d_v$, we encounter an inconsistent penalty along depth. Specifically, if the predicted wall is $d_v$ farther than the ground truth (total distance $d + d_v$), its IoU will be zero. Conversely, if the predicted wall is $d_v$ closer than the ground truth (total distance $d - d_v$), the IoU remains at 0.5. This occurs because all voxels behind the surface are filled with duplicated predictions.
  Similarly, when the predicted depth is $d - 2d_v$, the resulting IoU is $\frac{1}{3}$, and so forth.
  }
  \label{fig:iou-bug}
\end{figure*}

The misalignment between qualitative and quantitative results is caused by the inconsistent penalty along the depth direction.
A toy example in Fig.~\ref{fig:iou-bug} reveals several issues with the current metrics:

\begin{enumerate}
  \item If the model fills all areas behind the surface, it inconsistently penalizes depth predictions. The model can obtain a higher IoU by filling all areas behind the surface and predicting a closer depth. This thick surface issue is very common in dense models trained with visible masks or 2D supervision.
  \item If the predicted occupancy represents a thin surface, the penalty becomes overly strict. Even a deviation of just one voxel results in an IoU of zero.
  \item The visible mask only considers the visible area at the current moment, reducing occupancy prediction to a depth estimation task and overlooking the scene completion ability.
\end{enumerate}

\subsection{Mean IoU by Ray Casting}

\begin{figure}[t]
  \centering
  \subfloat[Simulating LiDAR]{\includegraphics[width=0.3\linewidth]{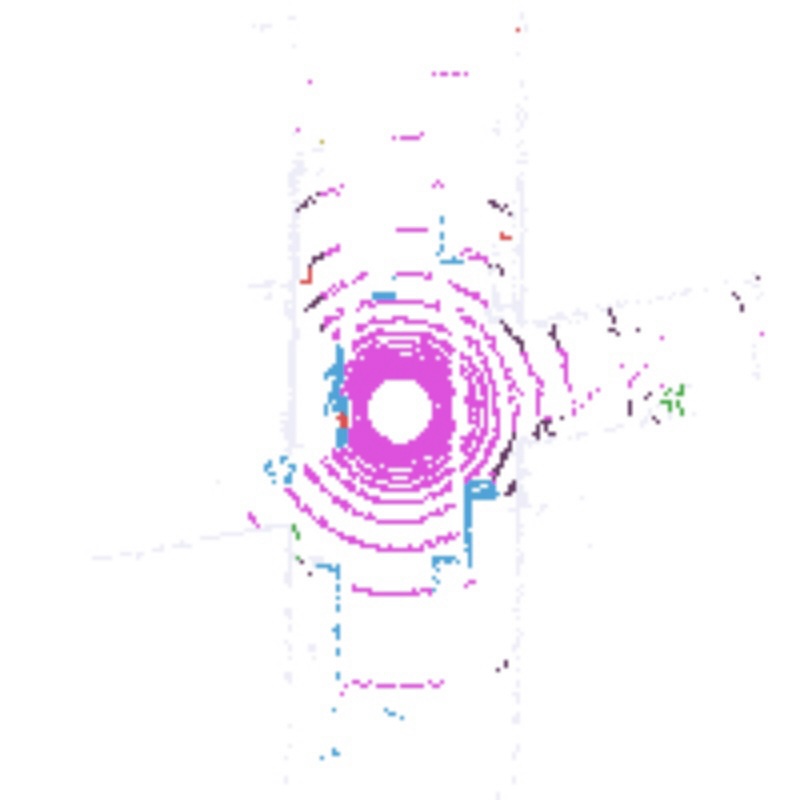}}%
  \hfill
  \subfloat[Equal-distant resampling]{\includegraphics[width=0.3\linewidth]{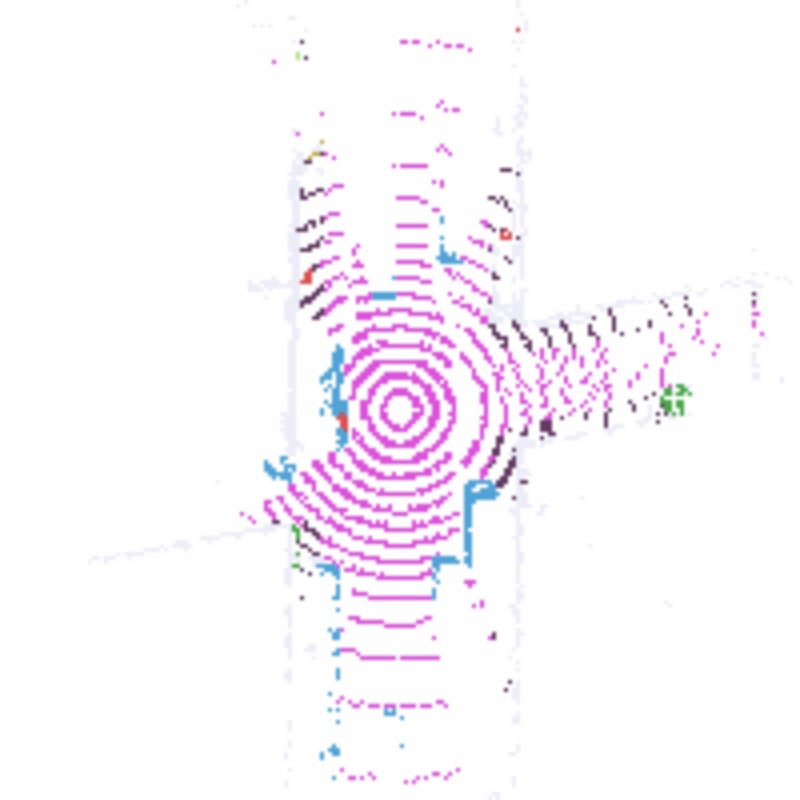}}%
  \hfill
  \subfloat[Temporal casting]{\includegraphics[width=0.3\linewidth]{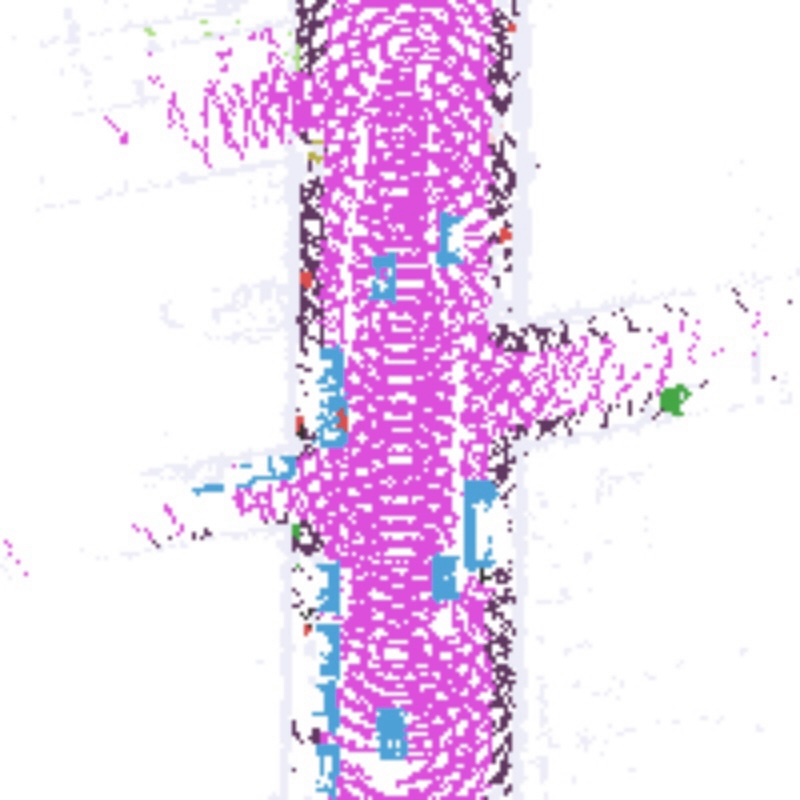}}%
  \caption{
  Covered area of RayIoU. 
  (a) The raw LiDAR ray samples are unbalanced at different distances.
  (b) We resample the rays to balance the weight on distance.
  (c) To investigate the performance of scene completion, we propose evaluating occupancy in the visible area on a wide time span, by casting rays on visited waypoints.
  }
  \label{fig:ray-cover}
\end{figure}

To address the above issues, we propose a new evaluation metric: Ray-level mIoU (RayIoU for short). 
In RayIoU, the set elements are query rays rather than voxels.
We emulate LiDAR behavior by projecting query rays into the predicted 3D occupancy volume. 
For each query ray, we compute the distance it travels before intersecting any surface and retrieve the corresponding class label.
We then apply the same procedure to the ground-truth occupancy to obtain the ground-truth depth and class label. 
In case a ray does not intersect with any voxel present in the ground truth, it will be excluded from the evaluation process.

As shown in Fig.~\ref{fig:ray-cover} (a), the raw LiDAR rays in a real dataset tend to be unbalanced from near to far.
Thus, we resample the rays to achieve a balanced distribution across different distances (Fig.~\ref{fig:ray-cover} (b)).
In the near field, we modify the ray channels to achieve equal-distant spacing when projected onto the ground plane.
In the far field, we increase the angular resolution of the ray channels to ensure a more uniform data density across varying ranges.
Moreover, our query ray can originate from the LiDAR position at the current, past, or future moments of the ego path.
Temporal casting (Fig.~\ref{fig:ray-cover} (c)) allows us to evaluate scene completion performance while maintaining a well-posed task. 
% The implementation details of ray casting are provided in supplementary material.

A query ray is classified as a \textit{true positive} (TP) if the class labels coincide and the L1 error between the ground-truth depth and the predicted depth is less than a certain threshold (\eg, 2m). Let $C$ be the number of classes, then RayIoU is calculated as follows:
\begin{equation}
  \text{RayIoU}=\frac{1}{C}\displaystyle \sum_{c=1}^{C}\frac{\text{TP}_c}{\text{TP}_c+\text{FP}_c+\text{FN}_c},
\end{equation}
where $\text{TP}_c$, $\text{FP}_c$ and $\text{FN}_c$ correspond to the number of true positive, false positive, and false negative predictions for class $c_i$.

RayIoU addresses all three of the aforementioned problems:

\begin{enumerate}
  \item Since the query ray calculates the distance to the first voxel it touches, the model cannot obtain a higher IoU by predicting a thicker surface.
  \item RayIoU determines true positives based on a distance threshold, which mitigates the overly strict nature of voxel-level mIoU.
  \item The query ray can originate from any position in the scene. This flexibility allows RayIoU to consider the model's scene completion ability, preventing the reduction of occupancy estimation to mere depth prediction.
\end{enumerate}

\section{Experiments}

We evaluate our model on the Occ3D-nuScenes \cite{occ3d} dataset. Occ3D-nuScenes is based on the nuScenes \cite{nuscenes} dataset, which consists of large-scale multimodal data collected from 6 surround-view cameras, 1 lidar and 5 radars. The dataset has 1000 videos in total and is split into 700/150/150 videos for training/validation/testing. Each video has roughly 20s duration and the key samples are annotated every 0.5s.

We use the proposed RayIoU to evaluate the semantic segmentation performance.
The query rays originate from 8 LiDAR positions of the ego path.
We calculate RayIoU under three distance thresholds: 1, 2 and 4 meters.
The final ranking metric is averaged over these distance thresholds.

\begin{table*}[t]
  \setlength{\tabcolsep}{5pt}
   \centering
   \caption{3D occupancy prediction performance on Occ3D-nuScenes \cite{occ3d}. We use RayIoU to compare our SparseOcc with other methods. ``8f'' and ``16f'' mean fusing temporal information from 8 or 16 frames. SparseOcc outperforms all existing methods under a weaker setting.}
   \label{table:occ3d-nus}
   \scalebox{0.91}{
   \begin{tabular}{l|ccc|c|ccc|c|c}
      \toprule
      Method & Backbone & Input Size & Epoch & \cellcolor[gray]{0.93}{RayIoU} & \multicolumn{3}{c|}{RayIoU\textsubscript{1m, 2m, 4m}} & mIoU & FPS \\
      \midrule
      BEVFormer (4f) \cite{bevformer} & R101   & 1600$\times$900 & 24 & \cellcolor[gray]{0.93}{32.4} & 26.1 & 32.9 & 38.0 & 39.2 & 3.0 \\
      RenderOcc \cite{renderocc} & Swin-B & 1408$\times$512 & 12 & \cellcolor[gray]{0.93}{19.5} & 13.4 & 19.6 & 25.5 & 24.4 & - \\
      SimpleOcc \cite{simpleocc} & R101 & 672$\times$336 & 12 & \cellcolor[gray]{0.93}{22.5} & 17.0 & 22.7 & 27.9 & 31.8 & 9.7 \\
      BEVDet-Occ (2f) \cite{bevdet4d} & R50    & 704$\times$256  & 90 & \cellcolor[gray]{0.93}{29.6} & 23.6 & 30.0 & 35.1 & 36.1 & 2.6 \\
      BEVDet-Occ-Long (8f)   & R50    & 704$\times$384  & 90 & \cellcolor[gray]{0.93}{32.6} & 26.6 & 33.1 & 38.2 & \textbf{39.3} & 0.8 \\
      FB-Occ (16f) \cite{fbocc}       & R50    & 704$\times$256  & 90 & \cellcolor[gray]{0.93}{33.5} & 26.7 & 34.1 & 39.7 & 39.1 & 10.3 \\
      \midrule
      SparseOcc (8f)                  & R50    & 704$\times$256  & 24 & \cellcolor[gray]{0.93}{34.0} & 28.0 & 34.7 & 39.4 & 30.1 & \textbf{17.3} \\
      SparseOcc (16f)                 & R50    & 704$\times$256  & 24 & \cellcolor[gray]{0.93}{35.1} & 29.1 & 35.8 & 40.3 & 30.6 & 12.5 \\
      SparseOcc (16f)                 & R50    & 704$\times$256  & 48 & \cellcolor[gray]{0.93}{\textbf{36.1}} & \textbf{30.2} & \textbf{36.8} & \textbf{41.2} & 30.9 & 12.5 \\
      \bottomrule
   \end{tabular}
   }
\end{table*}

\subsection{Implementation Details}

We implement our model using PyTorch \cite{pytorch}. Following previous methods, we adopt ResNet-50 \cite{resnet} as the image backbone. The mask transformer consists of 3 layers with shared weights across different layers. In our main experiments, we employ semantic queries where each query corresponds to a semantic class, rather than an instance. The ray casting module in RayIoU is implemented based on the codebase of \cite{pc_forcasting}.

During training, we use the AdamW \cite{adamw} optimizer with a global batch size of 8. The initial learning rate is set to $2 \times 10^{-4}$ and is decayed with cosine annealing policy. For all experiments, we train our models for 24 epochs. FPS is measured on a Tesla A100 GPU with the PyTorch \texttt{fp32} backend.

\subsection{Main Results}

In Tab. \ref{table:occ3d-nus} and Fig. \ref{fig:intro} (b), we compare SparseOcc with previous state-of-the-art methods on the validation split of Occ3D-nuScenes.
Despite under a weaker setting (ResNet-50 \cite{resnet}, 8 history frames, and input image resolution of 704 $\times$ 256), SparseOcc significantly outperforms previous methods including FB-Occ, the winner of CVPR 2023 occupancy challenge, with many complicated designs including forward-backward view transformation, depth net, joint depth and semantic pre-training, and so on. SparseOcc achieves better results (+1.6 RayIoU) while being much faster and simpler than FB-Occ, which demonstrates the superiority of our solution.

\begin{figure*}[t]
  \centering
  \subfloat{\includegraphics[width=1.0\linewidth]{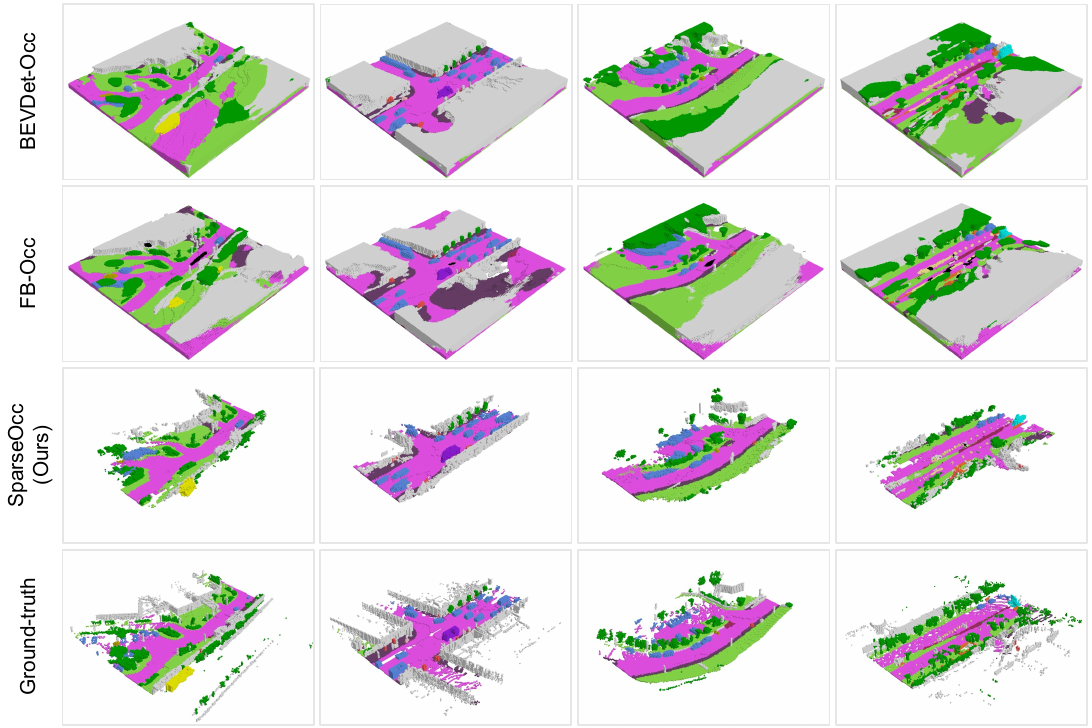}}
  \caption{Visualized comparison of semantic occupancy prediction. Despite discarding over 90\% of voxels, our SparseOcc effectively models the geometry of the scene and captures fine-grained details (\eg, the yellow-marked traffic cone in the bottom row).}
  \label{fig:semantic_viz}
\end{figure*}

We further provide qualitative results in Fig. \ref{fig:semantic_viz}. Both BEVDet-Occ and FB-Occ are dense methods and make many redundant predictions behind the surface. In contrast, SparseOcc discards over 90\% of voxels while still effectively modeling the geometry of the scene and capturing fine-grained details.

\subsection{Ablations}

In this section, we conduct ablations on the validation split of Occ3D-nuScenes to confirm the effectiveness of each module. By default, we use the single frame version of SparseOcc as the baseline. The choice for our model is made \textbf{bold}.

\begin{table*}[t]
  \setlength{\tabcolsep}{5pt}
  \centering
  \caption{Sparse voxel decoder vs. dense voxel decoder. Our sparse voxel decoder achieves nearly 4$\times$ faster inference speed than the dense counterparts.}
  \label{table:sparse-voxel-decoder}
  \scalebox{0.91}{
  \begin{tabular}{l|c|ccc|c}
    \toprule
    Voxel Decoder & RayIoU & RayIoU\textsubscript{1m} & RayIoU\textsubscript{2m} & RayIoU\textsubscript{4m} & FPS \\
    \midrule
    Dense coarse-to-fine & \textbf{29.9} & \textbf{24.0} & 30.4 & \textbf{35.4} & 6.3 \\
    Dense patch-based & 25.8 & 20.4 & 26.0 & 30.9 & 7.8 \\ 
    \midrule
    \textbf{Sparse coarse-to-fine} & \textbf{29.9} & 23.9 & \textbf{30.5} & 35.2 & \textbf{24.0} \\
    \bottomrule
  \end{tabular}
  }
\end{table*}

\paragraph{Sparse voxel decoder vs. dense voxel decoder.} In Tab. \ref{table:sparse-voxel-decoder}, we compare our sparse voxel decoder to the dense counterparts. Here, we implement two baselines, and both of them output a dense feature map with shape as 200$\times$200$\times$16$\times$$C$. The first baseline is a coarse-to-fine architecture without pruning empty voxels. In this baseline, we also replace self-attention with 3D convolution and use 3D deconvolution to upsample predictions. The other baseline is a patch-based architecture by dividing the 3D space into a small number of patches as PETRv2 \cite{petrv2} for BEV segmentation. We use 25$\times$25$\times$2 = 1250 queries and each one of them corresponds to a specific patch of shape 8$\times$8$\times$8. A stack of deconvolution layers are used to lift the coarse queries to a full-resolution 3D volume.

As we can see from the table, the dense coarse-to-fine baseline achieves a good performance of 29.9 RayIoU but with a slow inference speed of 6.3 FPS. The patch-based one is slightly faster with 7.8 FPS inference speed but with a severe performance drop by 4.1 RayIoU. Instead, our sparse voxel decoder produces sparse 3D features in the shape of $K \times C$ (where $K$ = 32000 $\ll$ 200$\times$200$\times$16), achieving an inference speed that is nearly 4$\times$ faster than the counterparts without compromising performance. This demonstrates the necessity and effectiveness of our sparse design.

\begin{table*}[t]
  \setlength{\tabcolsep}{5pt}
  \centering
  \caption{Ablation of mask transformer (MT) and the cross attention module in MT. Mask-guided sparse sampling is stronger and faster than the dense cross attention.}
  \label{table:mask-guided-sparse-sampling}
  \scalebox{0.91}{
  \begin{tabular}{l|l|c|ccc|c}
    \toprule
    MT & Cross Attention & RayIoU & RayIoU\textsubscript{1m} & RayIoU\textsubscript{2m} & RayIoU\textsubscript{4m} & FPS \\
    \midrule
    - & - & 27.0 & 20.3 & 27.5 & 33.1 & \textbf{29.0} \\
    $\surd$ & Dense cross attention & 28.7 & 22.9 & 29.3 & 33.8 & 16.2 \\ 
    \midrule
    $\surd$ & Sparse sampling & 25.8 & 20.5 & 26.2 & 30.8 & 24.0 \\
    $\surd$ & \textbf{+ Mask-guided} & \textbf{29.2} & \textbf{23.4} & \textbf{29.8} & \textbf{34.5} & 24.0 \\
    \bottomrule
  \end{tabular}
  }
\end{table*}

\paragraph{Mask Transformer.} In Tab. \ref{table:mask-guided-sparse-sampling}, we ablate the effectiveness of the mask transformer. The first row is a simple per-voxel baseline which directly predicts semantics from the sparse voxel decoder using a stack of MLPs. Introducing mask transformer with vanilla cross attention (as it is the common practice in MaskFormer and Mask3D) gives a performance boost of 1.7 RayIoU, but inevitably slows down the inference speed as it attends to all locations in an image. Therefore, to speed up the dense cross-attention pipeline, we adopt a sparse sampling mechanism which brings a 50\% reduction in inference time. By further introducing the predicted masks to guide the generation of sampling points, we finally achieve 29.2 RayIoU with 24 FPS.

\begin{figure}[t]
  \centering
  \subfloat[Pruning by top-$k$]{\includegraphics[height=0.24\linewidth]{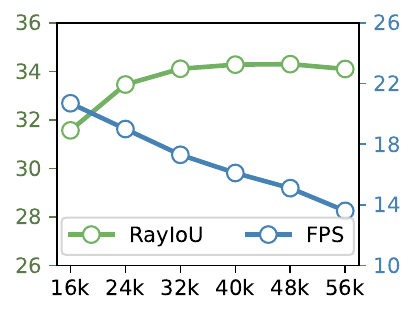}}%
  \hfill
  \subfloat[Pruning by thresholding]{\includegraphics[height=0.24\linewidth]{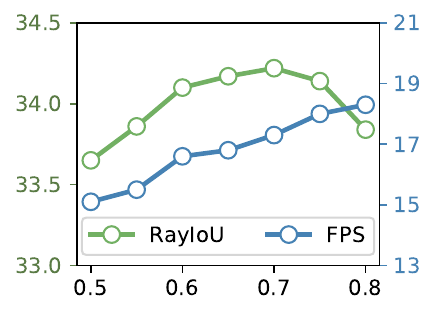}}%
  \hfill
  \subfloat[Number of frames]{\includegraphics[height=0.24\linewidth]{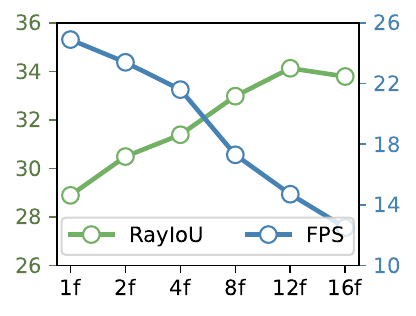}}%
  \caption{
  Ablations on voxel sparsity and temporal modeling.
  (a) The optimal performance occurs when $k$ is set to 32000 (5\% sparsity).
  (b) Top-$k$ can also be substituted with thresholding, \eg, voxels scoring less than a certain threshold will be pruned.
  (c) The performance continues to increase with the number of frames, but it starts to saturate after 12 frames.
  }
  \label{fig:frame_topk}
\end{figure}

\paragraph{Is a limited set of voxels sufficient to cover the scene?} In this study, we delve deeper into the impact of voxel sparsity on final performance. To investigate this, we systematically ablate the value of $k$ in Fig. \ref{fig:frame_topk} (a). Starting from a modest value of 16k, we observe that the optimal performance occurs when $k$ is set to 32k $\sim$ 48k, which is only 5\% $\sim$ 7.5\% of the total number of dense voxels (200$\times$200$\times$16 = 640000). Surprisingly, further increasing $k$ does not yield any performance improvements; instead, it introduces noise. Thus, our findings suggest that a $\sim$5\% sparsity level is sufficient. Keep increasing the density will reduce both accuracy and speed.

Pruning by top-$k$ is simple and effective, but it is related to specific dataset.
In real world, we can substitute top-$k$ with a thresholding method. Voxels scoring less than a given threshold (\eg, 0.7) will be pruned. Thresholding achieves similar performance to top-$k$ (see Fig. \ref{fig:frame_topk} (b)), and has the ability to generalize to different scenes.

\paragraph{Temporal modeling.} In Fig. \ref{fig:frame_topk} (c), we validate the effectiveness of temporal fusion. We can see that the temporal modeling of SparseOcc is very effective, with performance steadily increasing as the number of frames increases. The performance peaks at 12 frames and then saturates. However, the inference speed drops rapidly as the sampling points need to interact with every frame.

\subsection{More Studies}

\begin{table*}[t]
   \setlength{\tabcolsep}{0.006\linewidth}
   \centering
   \caption{To verify the effect of the visible mask, wo provide per-class RayIoU of BEVFormer and FB-Occ. $\dagger$ uses the visible mask during training. We find that training with the visible mask hurts the performance of background classes such as drivable surface, terrian and sidewalk.}
   \label{table:per-class-rayiou}
   \scalebox{0.81}{
   \begin{tabular}{l | c | c | c c c c c c c c c c c c c c c c c c}
       \toprule
       & & & \multicolumn{16}{c}{Per-class RayIoU} \\
       Method
       & \rotatebox{90}{{mIoU}}
       & \rotatebox{90}{RayIoU}
       & \rotatebox{90}{others} 
       & \rotatebox{90}{barrier} %
       & \rotatebox{90}{bicycle} %
       & \rotatebox{90}{bus} %
       & \rotatebox{90}{car} %
       & \rotatebox{90}{cons. veh.} %
       & \rotatebox{90}{motor.} %
       & \rotatebox{90}{pedes.} %
       & \rotatebox{90}{tfc. cone} %
       & \rotatebox{90}{trailer} %
       & \rotatebox{90}{truck} %
       & \rotatebox{90}{\cellcolor[gray]{0.93}{drv. surf.}} %
       & \rotatebox{90}{\cellcolor[gray]{0.93}{other flat}} %
       & \rotatebox{90}{\cellcolor[gray]{0.93}{sidewalk}} %
       & \rotatebox{90}{\cellcolor[gray]{0.93}{terrain}} %
       & \rotatebox{90}{manmade} %
       & \rotatebox{90}{vegetation} \\ %
       \midrule
       
       BEVFormer & {23.7} & \textbf{33.7}  & 5.0  & 42.2  & 18.2  & \textbf{55.2}  & \textbf{57.1}  & \textbf{22.7}  & 21.3  & 31.0  & \textbf{27.1}  & \textbf{30.7}  & 49.4  & \cellcolor[gray]{0.93}\textbf{58.4}  & \cellcolor[gray]{0.93}\textbf{30.4}  & \cellcolor[gray]{0.93}\textbf{29.4}  & \cellcolor[gray]{0.93}\textbf{31.7}  & 36.3  & 26.5 \\
       BEVFormer $\dagger$ & \textbf{{39.2}} & 32.4  & \textbf{6.4}  & \textbf{44.8}  & \textbf{24.0}  & \textbf{55.2}  & 56.7  & 21.0  & \textbf{29.8}  & \textbf{33.5}  & 26.8  & 27.9  & \textbf{49.5}  & \cellcolor[gray]{0.93}{45.8}  & \cellcolor[gray]{0.93}{18.7}  & \cellcolor[gray]{0.93}{22.4}  & \cellcolor[gray]{0.93}{18.5}  & \textbf{39.1}  & \textbf{29.8} \\
       \midrule
       FB-Occ & {27.9} & \textbf{35.6}  & \textbf{10.5}  & 44.8  & 25.6  & 55.6  & 51.7  & 22.6  & 27.2  & \textbf{34.3}  & \textbf{30.3}  & 23.7  & 44.1  & \cellcolor[gray]{0.93}\textbf{65.5}  & \cellcolor[gray]{0.93}\textbf{33.3}  & \cellcolor[gray]{0.93}\textbf{31.4}  & \cellcolor[gray]{0.93}\textbf{32.5}  & \textbf{39.6}  & \textbf{33.3} \\
       FB-Occ $\dagger$ & \textbf{{39.1}} & 33.5  & 5.0 & \textbf{44.9}  & \textbf{26.2}  & \textbf{59.7}  & \textbf{55.1}  & \textbf{27.9}  & \textbf{29.1}  & \textbf{34.3}  & 29.6  & \textbf{29.1}  & \textbf{50.5}  & \cellcolor[gray]{0.93}{44.4}  & \cellcolor[gray]{0.93}{22.4}  & \cellcolor[gray]{0.93}{21.5}  & \cellcolor[gray]{0.93}{19.5}  & 39.3  & 31.1 \\
   \bottomrule
   \end{tabular}
   }
\end{table*}

\begin{figure*}
  %\centering
  \includegraphics[width=\linewidth]{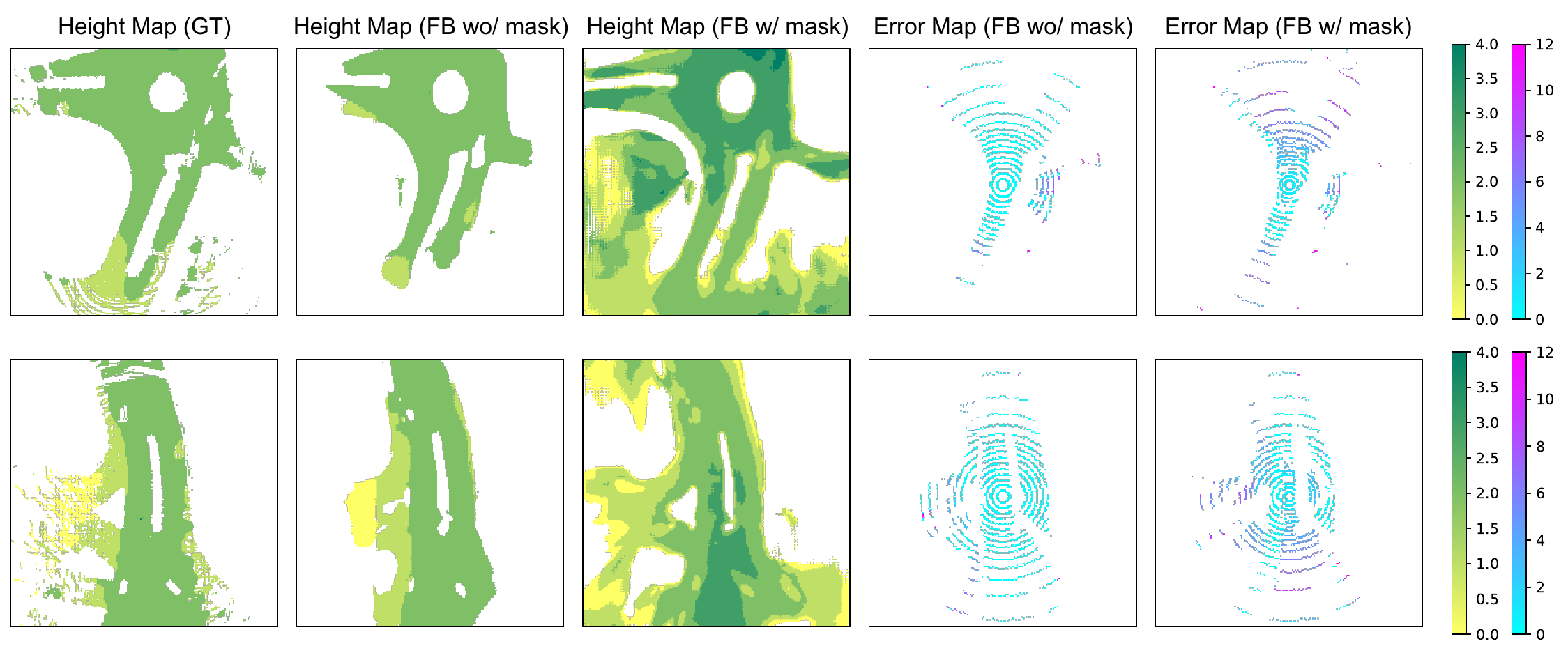}%
  \caption{Why does the performance of background classes, such as drivable surfaces, degrade when using the visible mask during training? We provide a visualization of the drivable surface as predicted by FB-Occ. Here, “FB w/ mask” and “FB wo/ mask” denote training with and without the visible mask, respectively. We observe that “FB w/ mask” tends to predict a higher and thicker road surface, resulting in significant depth errors along a ray. In contrast, “FB wo/ mask” predicts a road surface that is both accurate and consistent.}
  \label{fig:why-bg-worse}
\end{figure*}

\paragraph{The effect of training with visible masks.} Interestingly, we observed a peculiar phenomenon. Under the traditional voxel-level mIoU metric, dense methods can significantly benefit from disregarding the non-visible voxels during training. These non-visible voxels are indicated by a binary visible mask provided by the Occ3D-nuScenes dataset. However, we find that this strategy actually impairs performance under our new RayIoU metric. For instance, we train two variants of BEVFormer: one uses the visible mask during training, and the other does not. As shown in Tab. \ref{table:per-class-rayiou}, the former scores 15 points higher than the latter on the voxel-based mIoU, but it scores 1 point lower on RayIoU. This phenomenon is also observed on FB-Occ.

To further explore this, we present the per-class RayIoU in Tab. \ref{table:per-class-rayiou}. The table reveals that training with the visible mask enhances performance for most foreground classes such as bus, bicycle, and truck. However, it negatively impacts background classes like drivable surface and terrain.

This observation raises a further question: \textit{Why does the performance of the background category degrade?} To address this, we offer a visual comparison of the depth errors and height maps of the predicted drivable surface from FB-Occ in Fig. \ref{fig:why-bg-worse}, both with and without the use of visible mask during training. The figure illustrates that training with visible masks results in a thicker and higher ground prediction, leading to substantial depth errors in distant areas. Conversely, models trained without the visible mask predict depth with greater accuracy.

From these observations, we derive some valuable insights: ignoring non-visible voxels during training benefits foreground classes by resolving the issue of ambiguous labeling of unscanned voxels. However, it also compromises the accuracy of depth estimation, as models tend to predict a thicker and closer surface. We hope that our findings will benefit future research.

\paragraph{Panoptic occupancy.} We then show that SparseOcc can be easily extended for panoptic occupancy prediction, a task derived from panoptic segmentation that segments images to not only semantically meaningful regions but also to detect and distinguish individual instances. Compared to panoptic segmentation, panoptic occupancy prediction requires the model to have geometric awareness in order to construct the 3D scene for segmentation. By additionally introducing instance queries to the mask transformer, we seamlessly achieve the first fully sparse panoptic occupancy prediction framework using camera-only inputs.

Firstly, we utilize the ground-truth bounding boxes from the 3D object detection task to generate the panoptic occupancy ground truth. Specifically, we define eight instance categories (including \texttt{car}, \texttt{truck}, \texttt{construction} \texttt{vehicle}, \texttt{bus}, \texttt{trailer}, \texttt{motorcycle}, \texttt{bicycle}, \texttt{pedestrian}) and ten staff categories (including \texttt{terrain}, \texttt{manmade}, \texttt{vegetation}, etc). Each instance segment is identified by grouping the voxels inside the bounding box based on an existing semantic occupancy benchmark, such as Occ3D-nuScenes.

% However, we observe that using the original size of the box for grouping may cause some boundary voxels being missed due to the compactness of the box. Enlarging the box (such as 1.2x) leads to excessive overlap between boxes. To address these issues, we designed a two-stage grouping scheme. In the first stage, we use the original size of the box for grouping. Then, for voxels that have not been assigned to a specific instance, we select the closest box and assign it. This scheme effectively resolves the problems of boundary omission and box overlap.

We then design RayPQ based on the well-known panoptic quality (PQ) \cite{panoptic} metric, which is defined as the multiplication of \emph{segmentation quality} (SQ) and \emph{recognition quality} (RQ):

\begin{equation}
  {\text{PQ}} = \underbrace{\frac{\sum_{(p, g) \in \TP} \text{IoU}(p, g)}{\vphantom{\frac{1}{2}}|\TP|}}_{\text{segmentation quality (SQ)}} \times \underbrace{\frac{|\TP|}{|\TP| + \frac{1}{2} |\FP| + \frac{1}{2} |\FN|}}_{\text{recognition quality (RQ) }} \,,
\end{equation}
where the definition of true positive (TP) is the same as that in RayIoU. The threshold of IoU between prediction $p$ and ground-truth $g$ is set to 0.5.

\begin{table*}[t]
  \setlength{\tabcolsep}{5pt}
  \centering
  \caption{Panoptic occupancy prediction performance on Occ3D-nuScenes.}
  \label{table:pano}
  \scalebox{0.91}{
  \begin{tabular}{l|ccc|c|ccc}
     \toprule
     Method & Backbone & Input Size & Epoch & \cellcolor[gray]{0.93}{RayPQ} & RayPQ\textsubscript{1m} & RayPQ\textsubscript{2m} & RayPQ\textsubscript{4m} \\
     \midrule
     SparseOcc & R50 & 704$\times$256 & 24 & \cellcolor[gray]{0.93}{\textbf{14.1}} & 10.2 & 14.5 & 17.6 \\
     \bottomrule
  \end{tabular}
  }
\end{table*}

\begin{figure}[t]
  \centering
  \subfloat{\includegraphics[width=\linewidth]{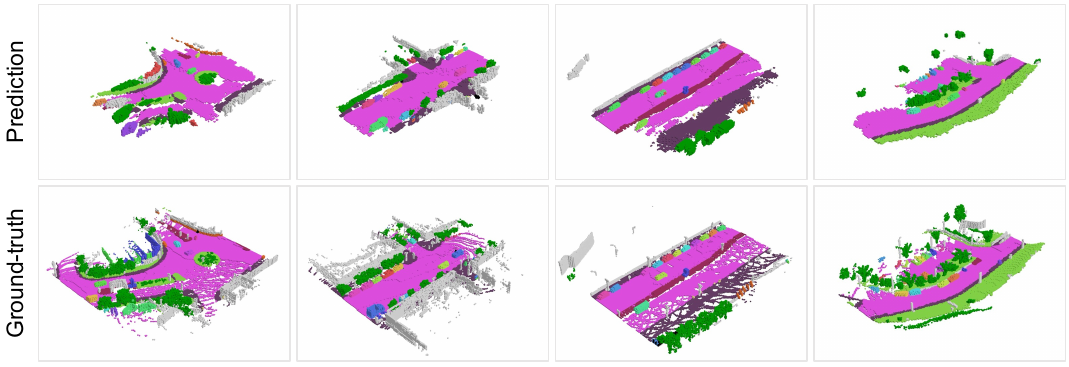}}
  \caption{Panoptic occupancy prediction. Different instances are distinguished by colors. Our model can capture fine-grained objects and road structures simultaneously.}
  \label{fig:pano_viz}
\end{figure}

In Tab. \ref{table:pano}, we report the performance of SparseOcc on panoptic occupancy benchmark. Similar to RayIoU, we calculate RayPQ under three distance thresholds: 1, 2 and 4 meters. SparseOcc achieves an averaged RayPQ of 14.1. The visualizations are presented in Fig. \ref{fig:pano_viz}.

% \paragraph{Enhancing sparsity by removing the road surface.} The majority of non-free occupancy data pertains to background geometry. In practice, the drivable surface occupancy can be effectively substituted with High-Definition Map (HD Map) or online mapping techniques~\cite{chen2022persformer,MapTR,wang2023openlanev2,li2023lanesegnet}. This replacement not only streamlines the sparsity but also enriches the semantic and structural understanding of roads. We construct experiments to investigate the effect of removing road surface in SparseOcc. The details can be found in the appendix.

\subsection{Limitations}

\paragraph{Accumulative errors.} In order to implement a fully sparse architecture, we discard a large number of empty voxels in the early stages. However, empty voxels that are mistakenly discarded cannot be recovered in subsequent stages. Moreover, the prediction of the mask transformer is constrained within a space predicted by the sparse voxel decoder. Some ground-truth instances do not appear in this unreliable space, leading to inadequate training of the mask transformer.

\section{Conclusion}

In this paper, we proposed a fully sparse occupancy network, named SparseOcc, which neither relies on dense 3D feature, nor has sparse-to-dense and global attention operations. 
We also created RayIoU, a ray-level metric for occupancy evaluation, eliminating the inconsistency flaws of previous metric.
Experiments show that SparseOcc achieves the state-of-the-art performance on the Occ3D-nuScenes dataset for both speed and accuracy. 
We hope this exciting result will attract more attention to the fully sparse 3D occupancy paradigm.

\iffalse
\subsection{Most Frequently Encountered Issues}
Please kindly use the checklist below to deal with some of the most frequently encountered issues in the latex files of ECCV submissions.

\begin{itemize}
\item I have removed all \verb| \vspace| and \verb|\hspace|  commands from my paper.
\item I have not used \verb|\cite| command in the abstract.
\item I have entered a correct \verb|\titlerunning{}| command and selected a meaningful short name for the paper.
\item I have used the same name spelling in all my papers accepted to ECCV and ECCV Workshops.
\item I have added acknowledgments without a section number, e.g. using the \verb|\section*{}| command.
\item Excluding references and acknowledgments, my paper is no longer than 14 pages.
\item I have not decreased the font size of any part of the paper (except tables) to fit into 14 pages, I understand Springer editors will remove such commands.
\end{itemize}
\fi

\section*{Acknowledgements}

We thank the anonymous reviewers for their suggestions that make this work better. This work is supported by the National Key R$\&$D Program of China (No. 2022ZD0160900), the National Natural Science Foundation of China (No. 62076119, No. 61921006), the Fundamental Research Funds for the Central Universities (No. 020214380119), and the Collaborative Innovation Center of Novel Software Technology and Industrialization.

\bibliographystyle{splncs04}
\bibliography{main}
\end{document}